\documentclass[runningheads]{llncs}
\usepackage{graphicx}

\usepackage{tikz}
\usepackage{comment}
\usepackage{amsmath,amssymb} 
\usepackage{color}

\usepackage[accsupp]{axessibility}

\usepackage{booktabs} 
\usepackage{cite}
\usepackage{bm} 
\usepackage{subfig}
\usepackage{diagbox}

\usepackage{float}  
\usepackage{multirow}
\usepackage{xcolor,colortbl} 
\usepackage{color}  
\usepackage[skip=2pt]{caption}  
\usepackage{selectp}  

\newcommand{\resnet}{ResNet}
\newcommand{\modelres}{$\Pi$-net-\resnet}
\newcommand{\sne}{SENet}
\newcommand{\noshare}{PDC}
\usepackage[switch]{lineno}

\usepackage{graphicx}
\usepackage{xcolor}

\newcommand{\tablestyle}[2]{\setlength{\tabcolsep}{#1}\renewcommand{\arraystretch}{#2}\centering\footnotesize}

\newcommand{\gr}{\rowcolor[gray]{.95}}

\newcommand{\realnum}{\mathbb{R}}
\providecommand{\naturalnum}{\mathbb{N}}
\providecommand{\bmcal}[1]{\bm{\mathcal{#1}}}

 \providecommand{\matnot}[1]{_{[{#1}]}}  
 \providecommand{\invar}{z}  
\providecommand{\binvar}{\bm{\invar}}  
\providecommand{\minvar}{\bm{Z}}  
\providecommand{\mvinvar}{\hat{\minvar}}  
\providecommand{\outvar}{x}  
\providecommand{\boutvar}{\bm{\outvar}}  
\providecommand{\matr}{\bm{\Phi}}  
\providecommand{\matra}[2]{\matr_{#1}^{[#2]}}  

\providecommand\eg{e.g.,}
\providecommand\ie{i.e.,}

\providecommand{\citep}{\cite} 
\providecommand{\citet}{\cite}

\usepackage[symbol]{footmisc}

\begin{document}
\pagestyle{headings}
\mainmatter
\def\ECCVSubNumber{8086}  

\title{Augmenting Deep Classifiers with Polynomial Neural Networks}

\titlerunning{Augmenting Deep Classifiers with Polynomial Neural Networks}
\author{Grigorios G. Chrysos*\inst{1} \and
Markos Georgopoulos*\inst{2} 
\and 
Jiankang Deng\inst{3} 
\and
Jean Kossaifi\inst{4} 
\and
Yannis Panagakis\inst{5} 
\and
Anima Anandkumar\inst{4} 
}
\authorrunning{Grigorios G. Chrysos et al.}
\institute{EPFL, Switzerland \and
Imperial College London, UK \and
Huawei, UK \and
NVIDIA, USA \and
University of Athens, Greece\\
\textsuperscript{*} First two authors have contributed equally.\\
\email{grigorios.chrysos@epfl.ch},
\email{m.georgopoulos@imperial.ac.uk}}

\maketitle

\begin{abstract}
Deep neural networks have been the driving force behind the success in classification tasks, e.g., object and audio recognition.
Impressive results and generalization have been achieved by a variety of recently proposed architectures, the majority of which are seemingly disconnected.
In this work, we cast the study of deep classifiers under a unifying framework. In particular, we express state-of-the-art architectures (e.g., residual and non-local networks) in the form of different degree polynomials of the input. Our framework provides insights on the inductive biases of each model and enables natural extensions building upon their polynomial nature. The efficacy of the proposed models is evaluated on standard image and audio classification benchmarks. The expressivity of the proposed models is highlighted both in terms of increased model performance as well as model compression. Lastly, the extensions allowed by this taxonomy showcase benefits in the presence of limited data and long-tailed data distributions. We expect this taxonomy to provide links between existing domain-specific architectures. The source code is available at \url{https://github.com/grigorisg9gr/polynomials-for-augmenting-NNs}.
\end{abstract}

\section{Introduction}
\label{sec:intro}

The unprecedented performance of AlexNet~\cite{krizhevsky2012imagenet} in ImageNet classification~\cite{russakovsky2015imagenet} led to the resurgence of research in the field of neural networks. 
Since then, an extensive corpus of papers has been devoted to improving classification performance by modifying the architecture of the neural network. 
However, only a handful (of seemingly disconnected) architectures, such as \resnet~\cite{he2015deep} or non-local neural networks~\cite{wang2018non}, have demonstrated impressive generalization across different tasks (\eg{} \cite{zhang2018self}), domains (\eg{} \cite{won2019toward}) and modalities (\eg{} \cite{kim2018bilinear}). This phenomenon can be attributed to the challenging nature of devising a network and the lack of understanding regarding the assumptions that come with its design, i.e., its inductive bias. 

\begin{figure}[t!]
  \centering
    \includegraphics[width=0.7\linewidth]{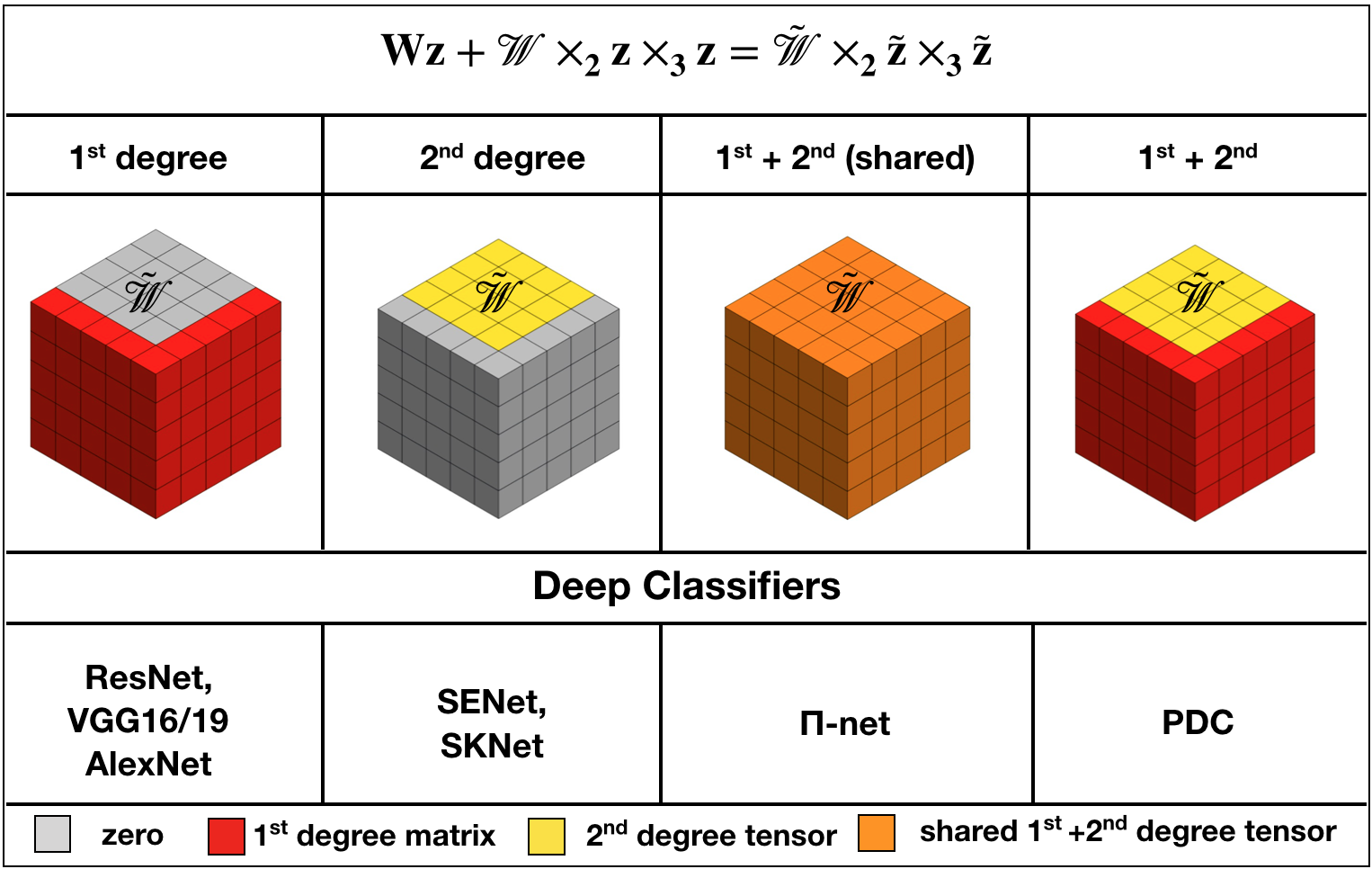}
  \caption{Parameter interactions for different degrees of polynomials. For visualization purpose we assume a second-degree expansion, where we have folded the learnable parameters into a tensor $\tilde{\bm{\mathcal{W}}}$ and the input is vectorized. The equation on the top is the second-degree polynomial in a tensor-format, while the $\tilde{\binvar}$ is a padded version of $\binvar$. Notice that first-degree polynomials~\cite{he2015deep, krizhevsky2012imagenet, simonyan2014very} have zeros in a large part of the tensor $\tilde{\bm{\mathcal{W}}}$, while similarly second-degree polynomials~\cite{hu2018squeeze,li2019selective} have the matrices connected to first-order interactions zero. On the contrary, $\Pi-$nets along with the proposed \noshare{} capture both interactions. Importantly, we illustrate how \noshare{} learns a more expressive model without the enforced sharing of the $\Pi-$net. The notation of mode-$m$ product along with the derivation is conducted in sec.~\ref{ssec:nosharing_explanation_motivational_figure} (supplementary).}
  \label{fig:graphical_abstract}
\end{figure}

Demystifying the success of deep neural architectures is of paramount importance. 
In particular, significant effort has been devoted to the study of neural architectures, \eg{} depth versus width of the neural network~\cite{rolnick2017power, hanin2019universal} and the effect of residual connections on the training of the network~\cite{hardt2016identity, huang2017densely, he2015deep}. In this work, we offer a principled approach to study state-of-the-art classifiers as polynomial expansions. We show that polynomials have been a recurring theme in numerous classifiers and interpret their design choices under a unifying framework.

The proposed framework provides a taxonomy for a collection of deep classifiers, \eg{} a non-local neural network is a third-degree polynomial and \resnet{} is a first-degree polynomial. 
Thus, we provide an intuitive way to study and extend existing networks as visualized in Fig.~\ref{fig:graphical_abstract}, as well as interpret their gap in performance. Lastly, we design extensions on existing methods and show that we can improve their classification accuracy or achieve parameter reduction. 
Concretely, our contributions are the following:
\begin{itemize}
    \item We express a collection of state-of-the-art neural architectures as polynomials. Our unifying framework sheds light on the inductive bias of each architecture. We experimentally verify the performance of different methods of the taxonomy on four standard benchmarks. 
    \item Our framework allows us to propose intuitive modifications on existing architectures. The proposed new architectures consistently improve upon their corresponding baselines, both in terms of accuracy as well as model compression.  
    \item We evaluate the performance under various changes in the training distribution, i.e., limiting the number of samples per class or creating a long-tailed distribution. The proposed models improve upon the baselines in both cases. 
    \item We release the code as open source to enable the reproduction of our results. 
\end{itemize}

 \section{Fundamentals on polynomial expansions}
\label{sec:nosharing_notation_and_intro_to_polynomials}
In this section, we provide the notation and an intuitive explanation on how a polynomial expansion emerges on various types of variables.

\noindent\textbf{Notation}: Below, 
we symbolize matrices (vectors) with bold capital (lower) letters, e.g. $\bm{X} (\bm{x})$. A variable that can be either a matrix or a vector is denoted by $\hat{\cdot}$. Tensors are considered as the multidimensional equivalent of matrices. Tensors are symbolized by calligraphic boldface letters e.g., $\bmcal{X}$. The symbol $\overrightarrow{\bm{1}}$ denotes a vector of ones and $\bmcal{I}$ is a third-order super-diagonal unit tensor. Due to constrained space, the detailed notation is on sec.~\ref{sec:nosharing_notation_supplem}, while in Table~\ref{tbl:nosharing_primary_symbols} the core symbols are summarized.

\begin{table}[]
    \caption{Symbols}
    \label{tbl:nosharing_primary_symbols}
    \centering
    \begin{tabular}{|c | c | c|}
    \toprule
    Symbol 	& Dimension(s) 		&	Definition \\
    \midrule
    $N$ 		            & $\naturalnum$		            &	Degree of polynomial expansion. \\
    $k$ 		            & $\naturalnum$		            & Rank of the decompositions. \\
    $\binvar$            & $\realnum^d$                      & Vector-form input to the expansion. \\
    $\odot, *$          &   -       & Khatri-Rao product, Hadamard product. \\
     \hline
    \end{tabular}
\end{table}

\noindent\textbf{Polynomial expansions}: Below, 
polynomials express a relationship between an input variable (\eg{} a scalar $\invar$) and (learnable) coefficients; this relationship only involves the two core operations of addition and multiplication. 
When the input variable is in vector form, \eg{} $\binvar \in \realnum^\delta$ with $\delta \in \naturalnum$, then the polynomial captures the relationships between the different elements of the input vector. 
The input variable can also be a higher-dimensional structure, \eg{} a matrix. This is frequently the case in computer vision, where one dimension can express spatial dimensions, while the other can express the features (channels). The polynomial can either capture the interactions across every element of the matrix with every  other element, or it can have higher-order interactions between specific elements, \eg{} the interactions of a row with each column.

\noindent\textbf{Relationship between polynomial expansions and tensors}: Multivariate polynomial expansions are intertwined with tensors. Specifically, the polynomial expansions of interest involve vector, matrix or tensor form as the input. 
Let us express the output $y \in \realnum$  as a $N^{\text{th}}$ degree polynomial expansion of a d-dimensional input $\binvar \in \realnum^d$:
\begin{align}
    \begin{split}
        y = \beta + \sum_{i=1}^d w^{[1]}_i \invar_i + \sum_{i=1}^d \sum_{j=1}^d w^{[2]}_{i, j} \invar_i \invar_j + \ldots + 
        \underbrace{\sum_{i=1}^d \cdots \sum_{j=1}^d}_{N \textit{ sums}} w^{[N]}_{i, \cdots, k} \invar_i \cdots \invar_k,
    \end{split}
    \label{eq:nosharing_elementwise_poly}
\end{align}
where $\beta \in \realnum$ is the constant term, and the $w$ terms are the scaling parameters for every degree. Notice that $w^{[n]}$ for $n > 2$ depends on $n$ indices, thus it can be expressed as a tensor. Collecting those tensors, $\{\bmcal{W}^{[n]} \in \realnum^{\underbrace{d \times \cdots \times d}_{n \textit{ times}}}\}_{n=1}^N$ represent learnable parameters in \eqref{eq:nosharing_elementwise_poly}. Due to the constrained space this relationship is further quantified in sec.~\ref{ssec:nosharing_explanation_motivational_figure}.

 \section{Polynomials and deep classifiers}
\label{sec:nosharing_taxonomy}

\begin{figure*}[t!]
  \centering
    \includegraphics[width=\linewidth]{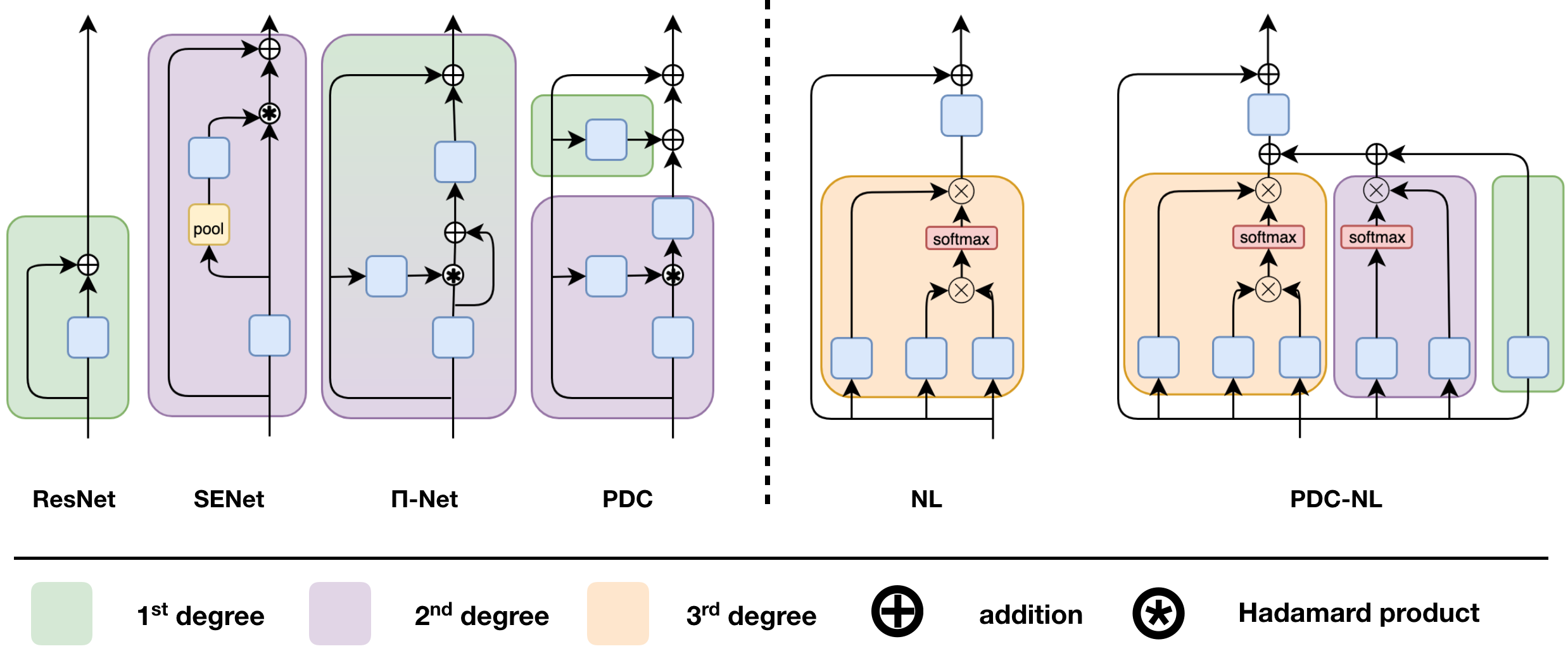}
    \caption{Blocks (up to third-degree) from different architectures. The layers (\ie{} blue boxes) denote any linear operation, \eg{} a convolution or a fully-connected layer, depending on the architecture. From left to right, the degree of the polynomial is increasing. Our framework enables also to complete the missing terms of the polynomial (\ie{} \noshare-NL versus NL). }
  \label{fig:nosharing_blocks}
\end{figure*}

The proposed framework unifies recent deep classifiers through the lens of polynomials. 
We formalize the framework of polynomials below. Formally, let functions $\matra{i}{d} (\mvinvar)$ define a linear (or multi-linear) function over $\mvinvar$. The input variable  $\mvinvar$ can either be a vector or a matrix, while $d$ declares the degree and $i$ the index of the function. Then, a polynomial of degree$-N$ is expressed as: 
\begin{equation}
\begin{split}
    \hat{\bm{Y}} = \hat{\bm{\beta}} + \matra{1}{1} (\mvinvar) + \matra{1}{2} (\mvinvar) \matra{2}{2} (\mvinvar) + \ldots + \underbrace{\matra{1}{N} (\mvinvar) \ldots \matra{N-1}{N}\matra{N}{N} (\mvinvar)}_\text{N terms}
\end{split}
\label{eq:nosharing_general_poly}
\end{equation}
where $\hat{\bm{\beta}}$ is the constant term. Evidently, each additional degree introduces new parameters that can grow up exponentially with respect to the degree. However, we posit that this introduces a needed inductive bias on the method. In addition, if reduced parameters are required, this can be achieved by using low-rank assumptions or by sharing parameters across different degrees.

Using the formulation in \eqref{eq:nosharing_general_poly}, we exhibit how well-established methods can be formulated as polynomials. We present a taxonomy of classifiers based on the degree of the polynomial. In particular, we present first, second, third and higher-degree polynomial expansions. For modeling purposes, we focus on the core block of each architecture, while ignoring any activation functions.

\noindent\textbf{First-degree polynomials} include the majority of the feed-forward networks, such as AlexNet~\cite{krizhevsky2012imagenet}, VGG~\cite{simonyan2014very}. Specifically, the networks that include stacked linear operations (fully-connected or convolutional layers) but do not include any matrix or elementwise multiplication of the representations fit in this category. Such networks can be expressed in the form $\hat{\bm{Y}} = \bm{C}\mvinvar + \hat{\bm{\beta}}$, where the weight matrix $\bm{C}$ is a learnable parameter. A special case is \resnet~\cite{he2015deep}. The idea is to introduce shortcut connections that enable residual learning. Notation-wise this is a re-parametrization of the weight matrix $\bm{C}$ as $\hat{\bm{Y}}_r = (\bm{I} + \bm{C})\mvinvar + \hat{\bm{\beta}}$ where $\bm{I}$ is an identity matrix. Thus, $\matra{1}{1}(\mvinvar) = (\bm{I} + \bm{C})\mvinvar$.

\noindent\textbf{Second-degree polynomials} model self-interactions, \ie{} they can selectively maximize the related inputs through second-order interactions. 
Often the interactions emerge in a particular dimension, \eg{} correlations of the channels only, based on the particular application. 

One special case of the second-degree polynomial is the \emph{Squeeze-and-excitation networks (\sne)}~\cite{hu2018squeeze}. The motivation lies in improving the channel-wise interactions, since there is already a strong inductive bias for the spatial interactions through convolutional layers. Notation-wise, the squeeze-and-excitation block is expressed as:
\begin{equation}
    {\bm{Y}}_{s} = (\minvar\bm{C}_1) * r(p(\minvar\bm{C}_1)\bm{C}_2),
    \label{eq:nosharing_senet_block}
\end{equation}
where $*$ denotes the Hadamard product, $p$ is the global pooling function, $r$ is a function that replicates the channels in the spatial dimensions and $\bm{C}_1, \bm{C}_2$ are weight matrices with learnable parameters. For simplicity, we assume the input is a matrix instead of a tensor, \ie{} $\minvar \in \realnum^{hw \times c}$ with $h$ the height, $w$ the width and $c$ the channels of the image. 
Then, ${\bm{Y}}_{s}$ expresses a second-degree term with $\matra{1}{2}(\minvar) = \minvar\bm{C}_1$ and $\matra{2}{2}(\minvar) = \frac{1}{hw} \bm{\mathcal{I}} \times_3 (\bm{C}_2^T\bm{C}_1^T \minvar^T \overrightarrow{\bm{1}})$ where $\bm{\mathcal{I}}$ is a third-order super-diagonal unit tensor and $\overrightarrow{\bm{1}}$ is a vector of ones.

The Squeeze-and-excitation block has been extended in the literature. The selective kernels networks~\cite{li2019selective} introduce a variant, where $\minvar$ is replaced with the transformed $\minvar \bm{U}_1 + \minvar\bm{U}_2$. The learnable parameters $\bm{U}_1, \bm{U}_2$ include different receptive fields. However, the degree of the polynomial remains the same, \ie{} second-degree.

The \emph{Factorized Bilinear Model}~\cite{li2017factorized} aim to extend the linear transformations of a convolutional layer by modeling the pairwise feature interactions. In particular, every output $y_{fb} \in \realnum$ is expressed as the following expansion of the input $\binvar \in \realnum^d$:
\begin{equation}
    y_{fb} = c_1 + \bm{c}_2^T\binvar + \binvar^T \bm{C}_3^T \bm{C}_3 \binvar,
    \label{eq:nosharing_factorized_bilinear_block}
\end{equation}
where $c_1 \in \realnum, \bm{c}_2 \in \realnum^d, \bm{C}_3 \in \realnum^{k\times d}$ are learnable parameters with $k \in \naturalnum$.

Arguably, a more general form of second-degree polynomial expansion is proposed in \emph{SORT}~\cite{wang2017sort}. The idea is to combine different branches with multiplicative interactions. SORT is expressed as: 
\begin{equation}
    {\bm{Y}}_{t} = \bm{C}_1\mvinvar + \bm{C}_2\mvinvar +  g((\bm{C}_1\mvinvar) * (\bm{C}_2\mvinvar)),
    \label{eq:nosharing_sort_block}
\end{equation}
where the $\bm{C}_1, \bm{C}_2$ are learnable parameters and $g$ an elementwise, differentiable function. When $\bm{C}_1$ is identity, they $g$ is the elementwise square root function.

\noindent\textbf{Third-degree polynomials} can encode long-range dependencies that are not captured by local operators, such as convolutions. 
A popular framework in this category is the \emph{non-local neural networks (NL)}~\cite{wang2018non}. The non-local block can be expressed as: 
\begin{equation}
    {\bm{Y}}_{n} = (\minvar \bm{C}_1 \bm{C}_2 \minvar^T)\minvar\bm{C}_3,
    \label{eq:nosharing_nl_block}
\end{equation}
where the matrices $\bm{C}_i$ for $i=1,2,3$ are learnable parameters and $\minvar \in \realnum^{hw \times c}$ is the input. The third-degree term is then $\matra{1}{3}(\minvar) = \minvar \bm{C}_1$,  $\matra{2}{3}(\minvar) = \bm{C}_2 \minvar^T$ and $\matra{3}{3}(\minvar) = \minvar\bm{C}_3$. 
Recently, \emph{disentangled non-local networks (DNL)}~\cite{yin2020disentangled} extends the formulation by including a second-degree term and the generated output is:
\begin{equation}
    {\bm{Y}}_{dn} = ((\minvar \bm{C}_1 - \bm{\mu}_q) (\bm{C}_2 \minvar - \bm{\mu}_k)^T + \minvar \bm{c}_4 \overrightarrow{\bm{1}}^T)\minvar\bm{C}_3,
    \label{eq:nosharing_dnl_block}
\end{equation}
where $\bm{c}_4$ is a weight vector, $\bm{\mu}_q, \bm{\mu}_k$ are the mean vectors of the keys and queries representations and $\overrightarrow{\bm{1}} \in \realnum^{hw\times 1}$ a vector of ones. This translates to a new second-degree term with $\matra{1}{2}(\minvar) = \minvar \bm{c}_4 \overrightarrow{\bm{1}}^T$ and $\matra{2}{2}(\minvar) = \minvar\bm{C}_3$.

\noindent\textbf{Higher-degree polynomials} can (virtually) approximate any smooth functions. The Weierstrass theorem~\cite{stone1948generalized} and its extension~\cite{nikol2013analysis} (pg 19) guarantee that any smooth function can be approximated by a higher-degree polynomial.

A recently proposed framework that leverages high-degree polynomials to approximate functions is ${\Pi}$-net~\cite{chrysos2020poly}. Each $\Pi-$net block is a polynomial expansion with a pre-determined degree. The learnable coefficients of the polynomial are represented with higher-order tensors. One of the drawbacks of this method is that the order of the tensor increases linearly with the degree, hence the parameters explode exponentially. To mitigate this issue, a coupled tensor decomposition that allows for sharing among the coefficients is utilized. Thus, the number of parameters is reduced significantly. Although the method is formulated as a complete polynomial, the proposed sharing of the coefficients suppresses its expressive power in favour of model compression.
The model used for the classification can be expressed with the following recursive relationship:

\begin{equation}
    \boutvar_{n} = \Big(\bm{A}\matnot{n}^T\binvar\Big) * \Big(\bm{S}\matnot{n}^T \boutvar_{n-1} + \bm{B}\matnot{n}^T\bm{b}\matnot{n}\Big) +  \boutvar_{n-1}
    \label{eq:prodpoly_model3}
\end{equation}
for an $N^{th}-$degree expansion order with $n=2,\ldots,N$. The weight matrices $\bm{A}\matnot{n}, \bm{S}\matnot{n},\bm{B}\matnot{n}$ and the weight vector $\bm{b}$ are learnable parameters, while $\boutvar_{1} = \Big(\bm{A}\matnot{1}^T\binvar\Big) * \Big( \bm{B}\matnot{1}^T\bm{b}\matnot{1} \Big)$. The recursive equation can be used to express an arbitrary degree of expansion, while the weight matrices are shared across different degree terms. For instance, $\bm{A}\matnot{2}$ is shared by both first and second-degree terms when $N=2$.

\noindent\textbf{From blocks to architecture:} The core blocks of different architectures, and their polynomial counterparts, are analyzed above. The final network (in each case) is obtained by concatenating the respective blocks in a cascade. That is, the output of the first block is used as the input for the next block and so on. Each block expresses a polynomial expansion, thus, the final architecture expresses a product of polynomials.

\section{Novel architectures based on the taxonomy}
\label{sec:nosharing_new_architectures}
The taxonomy offers a new perspective on how to modify existing architectures in a principled way. We showcase how new architectures arise in a natural way by modifying the popular Non-local neural network and the recent $\Pi-$nets.

\subsection{Higher-degree \resnet{} blocks}
\label{ssec:nosharing_pi_net_nosharing}
As shown in the previous section, the \resnet{} block is a first-degree polynomial. We can extend the polynomial expansion degree in order to enable higher-order correlations. 
A general $N^{th}-$degree polynomial is expressed as: 

\begin{equation}
    \bm{y} = \sum_{n=1}^N \bigg(\bmcal{W}^{[n]} \prod_{j=2}^{n+1} \times_{j} \binvar\bigg) + \bm{\beta}
    \label{eq:prodpoly_poly_general_eq}
\end{equation}
where $\binvar \in \realnum^{\delta}$, $\times_{m}$ denotes the mode-m vector product, $\big\{\bmcal{W}^{[n]} \in  \realnum^{o\times \prod_{m=1}^{n}\times_m \delta}\big\}_{n=1}^N$ are the tensor parameters. To reduce the learnable parameters, we assume a low-rank CP decomposition~\cite{kolda2009tensor} on each tensor. By applying Lemma~\ref{lemma:polygan_lemma_hadamard_kr2}, we obtain: 
\begin{equation}
\begin{split}
    \bm{y} = \bm{\beta} + \bm{C}_{1,[1]}^T \binvar + \Big(\bm{C}_{1,[2]}^T \binvar\Big) * \Big(\bm{C}_{2,[2]}^T \binvar\Big) + \ldots +  \underbrace{\Big(\bm{C}_{1,[N]}^T \binvar\Big) * \ldots * \Big(\bm{C}_{N,[N]}^T \binvar\Big)}_{\text{N Hadamard products}}
\end{split}
\label{eq:nosharing_model_no_sharing}
\end{equation}
where all $\bm{C}_i$ are learnable parameters. Our proposed model in \eqref{eq:nosharing_model_no_sharing} is modular and can be designed for arbitrary polynomial degree. A schematic of \eqref{eq:nosharing_model_no_sharing} is depicted in Fig~\ref{fig:nosharing_higher_order_block}. The proposed model differentiates itself from \eqref{eq:prodpoly_model3} by not assuming any parameter sharing across the different degree terms.

\subsection{Polynomial non-local blocks of different degrees}
\label{ssec:nosharing_nonlocal_different_degrees}

In this section, we demonstrate how the proposed taxonomy can be used to design a new architecture based on non-local blocks (NL). 
NL includes a third-degree term, while disentangled non-local network (DNL) includes both a third-degree and a second-degree term. 

We begin by including an additional first-degree term with learnable weights. The formed \noshare-$NL^{[3]}$ block is:
\begin{equation}
    {\bm{Y}}^{[3]}_{ours} = (\minvar \bm{C}_1 \bm{C}_2 \minvar^T)\minvar\bm{C}_3 + \minvar \bm{C}_4\minvar\bm{C}_5 + \minvar\bm{C}_6
    \label{eq:nosharing_complete_poly_nonlocal}
\end{equation}
where $\bm{C}_i$ are learnable parameters. In practice, a softmax activation function is added for the second and third degree factors, similar to the baseline. Besides the first-degree term, the proposed model differentiates itself from DNL by removing the sharing between the factors of third and second degree (i.e., $\bm{C}_3$ and $\bm{C}_5$) as well as utilizing a full factor matrix instead of a vector for the latter (i.e., $\bm{C}_4$).
In our implementation the matrices $\bm{C}_i$ matrices (for $i\neq4$) compress the channels of the input by a factor of $4$ for all models. 

Building on \eqref{eq:nosharing_complete_poly_nonlocal}, we propose to expand the \noshare-$NL^{[3]}$ to a fourth degree polynomial expansion. We multiply the right hand side of \eqref{eq:nosharing_complete_poly_nonlocal} with a linear function with respect to the input $\minvar$. That is, the \noshare-$NL^{[4]}$ block is:
\begin{equation}
    {\bm{Y}}^{[4]}_{ours} = {\bm{Y}}^{[3]}_{ours} + {\bm{Y}}^{[3]}_{ours} * r(p(\minvar\bm{C}_7)\bm{C}_8),
    \label{eq:nosharing_fourth_degree_nonlocal}
\end{equation}
where the term $r(p(\minvar\bm{C}_7)\bm{C}_8)$ is similar to squeeze-and-excitation right hand side term. The new term captures the channel-wise correlations.

 \section{Experimental evaluation}
\label{sec:nosharing_experiments}

In this section, we study how the degree of the polynomial expansion affects the expressive power of the model.
Our goal is to illustrate how the taxonomy enables improvements in strong-performing models with minimal and intuitive modifications. This approach should also shed light on the inductive bias of each model, as well as the reasons behind the gap in performance and model compression.
Unless mentioned otherwise, the degree of the polynomial for each block is considered the second-degree expansion. The proposed variants are referred to as \noshare. 
Experiments with higher-degree polynomials, which are denoted as \noshare$^{[k]}$ for $k^{th}$ degree, are also conducted.

\noindent\textbf{Training details:} For fair comparison all the aforementioned experiments have the same optimization-related hyper-parameters, \eg{} the layer initialization. The training is run for $120$ epochs with batch size $128$ and the SGD optimizer is used. The initial learning rate is $0.1$, while the learning rate is multiplied with a factor of $0.1$ in epochs $40, 60, 80, 100$. 
Classification accuracy is utilized as the evaluation metric.

\subsection{Image classification with residual blocks}
\label{ssec:nosharing_experiments_image_resnet}

The standard benchmarks of CIFAR10~\cite{krizhevsky2014cifar}, CIFAR100~\cite{CIFAR100} and ImageNet~\citep{russakovsky2015imagenet} are utilized to evaluate our framework on image classification. To highlight the impact of the degree of the polynomial, we experiment with first (\resnet), second (\sne{} and $\Pi-$net\footnote{The default $\Pi-$net block is designed as second-degree polynomial. Higher-degree blocks are denoted with the symbol $\cdot^{[k]}$.}) as well as higher-degree polynomials. \noshare{} relies on eq.~\ref{eq:nosharing_model_no_sharing}; that is, we do not assume shared factor matrices across the different degree terms.

\begin{figure}[t!]
  \centering
    \includegraphics[width=0.6\linewidth]{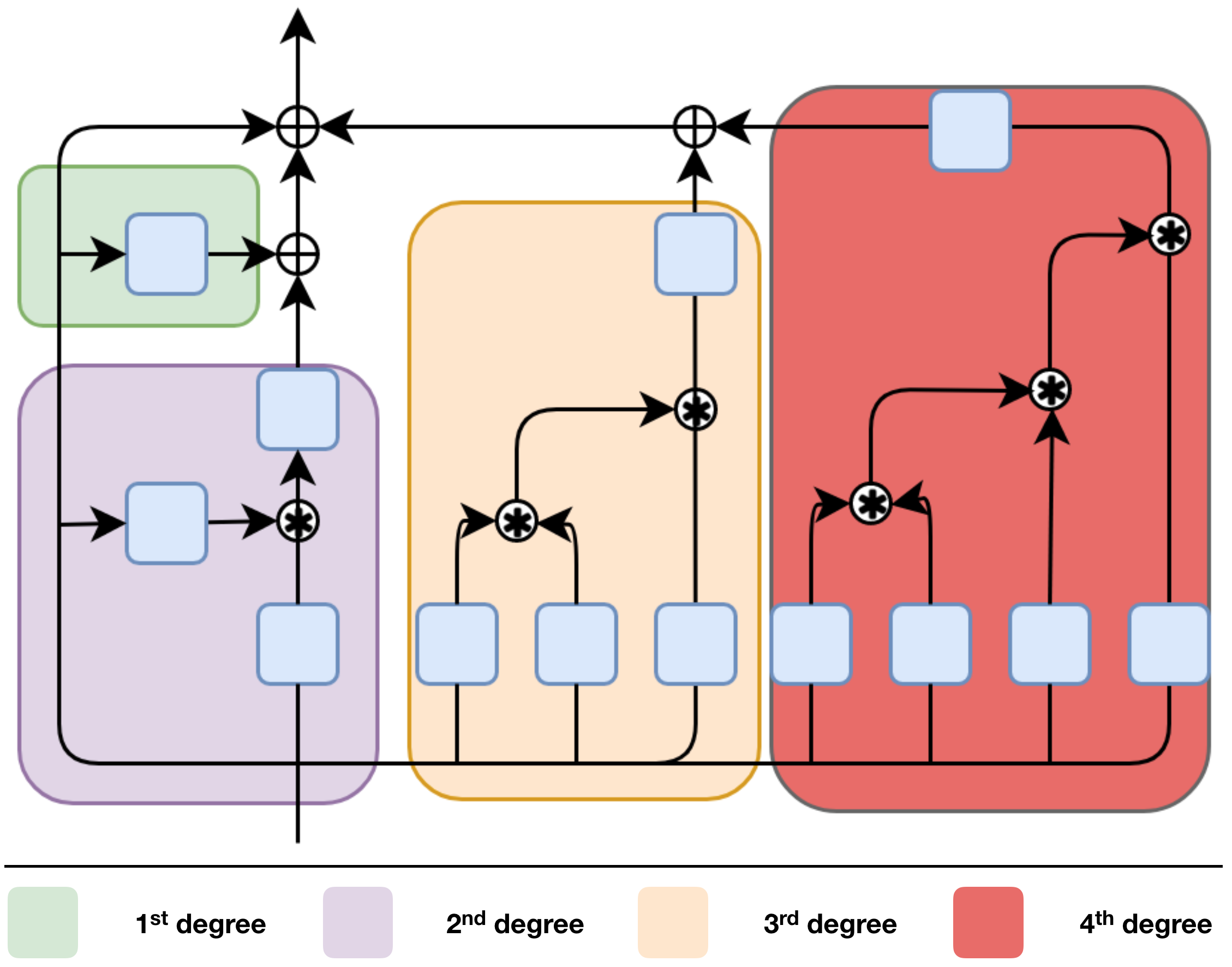}
    \caption{Illustration of the modular nature of the propose polynomial expansion. Notice how we can trivially add new blocks in \noshare{} to reach the pre-determined degree of the complete polynomial. The $N^{th}$-degree term includes $N+1$ new layers (\ie{} blue boxes), which might make it impractical to expand over the fourth-degree. In practice, higher-degree polynomials make the model more expressive and hence allow for reduced number of channels.}

  \label{fig:nosharing_higher_order_block}
\end{figure}

The first experiment is conducted on CIFAR100. The \resnet18 and the respective \sne{} are the baselines, while the \modelres{} is the main comparison method. To exhibit the efficacy of \noshare, we implement various versions: (a) \noshare-channels is the variant that retains the same channels as the original \resnet18, (b) \noshare-param has approximate the same parameters as the corresponding baselines, (c) \noshare-comp has the least parameters that can achieve a performance similar to the \resnet18 (which is achieved by reducing the channels), (d) the \noshare{} which is the variant that includes both reduced parameters and increased performance with respect to \resnet18, (e) the higher-degree variants of \noshare$^{[3]}$ and \noshare$^{[4]}$  that modify the \noshare{} such that the degree of each respective block is three and four respectively (see Fig.~\ref{fig:nosharing_higher_order_block}). 

The accuracy of each method is reported in Table~\ref{tab:nosharing_resnet_cifar100}. \sne{} improves upon the baseline of \resnet18, which verifies the benefit of second-degree correlations in the channels. This is further demonstrated with \modelres{} that captures second-degree information on all dimensions and can achieve the same performance with reduced parameters. Our method can further reduce the parameters by 60\% over \resnet18 and achieve the same accuracy. This can be attributed to the no-sharing scheme that enables more flexibility in the learned decompositions. The \noshare-param further improves the accuracy by increasing the parameters, while the \noshare-channels achieves the best accuracy by maintaining the same channels per residual block as \resnet18. The experiment is repeated with \resnet34 as the baseline. The accuracies in Table~\ref{tab:nosharing_resnet_cifar100_resnet34} demonstrate similar pattern, \ie{} the parameters can be substantially reduced without sacrificing the accuracy. The results exhibit that the expressive power of \noshare{} can allow it to be used both for compression and for improving the performance by maintaining the same number of parameters.

\begin{table}[h]
\centering
    \begin{minipage}{.48\linewidth}
    \caption{Image classification on CIFAR100 with variants of \resnet18. The symbol `p' denotes parameters.}  
     \begin{tabular}{|c | c | c|}
         \hline
         \textbf{Model} & \textbf{\#p} $\times 10^6$ & \textbf{Accuracy}\\
        \hline
         \resnet18 & $11.2$ & $0.756$\\\hline
         \sne       & $11.6$ & $0.760$ \\\hline
         \modelres & ${6.1}$ & $0.760$\\\hline
         \noshare-comp & $\bm{4.3}$ & $0.760$\\\hline
         \noshare-channels & $19.2$ & $\bm{0.773}$\\\hline  
         \noshare-param & $11.4$ & ${0.770}$\\\hline  
         \noshare & $8.0$ & ${0.765}$\\\hline
         \noshare$^{[3]}$ & $16.8$ & ${0.766}$\\\hline
         \noshare$^{[4]}$ & $28.0$ & ${0.771}$\\\hline 
     \end{tabular}
     \label{tab:nosharing_resnet_cifar100}
    \end{minipage}
    \begin{minipage}{.48\linewidth}
        \caption{Image classification on CIFAR10. The symbol `p' denotes parameters.}
        \begin{tabular}{|c | c | c|}
             \hline
             \textbf{Model} & \textbf{\#p} $\times 10^6$ & \textbf{Accuracy}\\
            \hline
             \resnet18 & $11.2$ & $0.945$\\\hline
             \sne       & $11.5$ & $0.946$ \\\hline
             \modelres & ${6.0}$ & $0.945$\\\hline
             \noshare & $8.0$ & $\bm{0.946}$\\\hline
             \noshare-comp & $\bm{4.3}$ & ${0.945}$\\\hline
             \hline\hline
             \resnet34 & $21.3$ & $0.948$\\
             \hline
             \modelres &  ${13.0}$ & $\bm{0.949}$\\\hline
             \noshare &  $\bm{10.5}$ & $0.948$\\\hline
         \end{tabular}
         \label{tab:nosharing_resnet_cifar10}
    \end{minipage}

\end{table}

A similar experiment is conducted on CIFAR10. 
The same networks as above are used as baselines, \ie{} \resnet18 and \sne{}. The methods \modelres{} and the \noshare{} variants are also implemented as above. The results in Table~\ref{tab:nosharing_resnet_cifar10} verify that both the \modelres{} and \noshare-comp can compress substantially the number of parameters required while retaining the same accuracy. Importantly, our method achieves a  parameter reduction of $28\%$ over \modelres.

The last experiment is conducted on ImageNet~\cite{deng2009imagenet}, which contains 1.28M training images and 50K validation images from 1000 classes. We follow the standard settings to train 
deep networks on the training set and report the single-crop top-1 and the top-5 errors on the validation set. Our pre-processing and augmentation strategy follows the settings of the baseline (i.e., \resnet18). All models are trained for 100 epochs on 8  
GPUs with 32 images per GPU (effective batch size of 256) with synchronous SGD of momentum 0.9. The learning rate is initialized to 0.1, and decays by a factor of 10 at the $30^{th}$, $60^{th}$, and $90^{th}$ epochs. The results in Table~\ref{tab:nosharing_resnet_imagenet} reflect the patterns that emerged above. The proposed \noshare{} can reduce the number of parameters  required while retaining the same accuracy. The variant of \noshare{} that has similar number of parameters to the baseline \resnet18 can achieve an increase in the accuracy. More specifically, we set the stem channel number to $52$ for PDC-ResNet18-cmp and $60$ for \noshare18, respectively. In addition, we also decrease the channel to $1/4$ before the Hadamard product in the proposed PDC block to save the computation cost.

\begin{table*}[h]
\caption{Image classification on ImageNet with variants of \resnet18.}
\centering
    \begin{tabular}{|c | c | c| c | c |}
    \hline
    Model   & Top-1 Accuracy & Top-5 Accuracy & Flops  & \textbf{\#p} $\times 10^6$\\
    \hline
        ResNet18        & 0.698 &  0.891 & 1.82G & 11.69M \\
        SE-ResNet18     & 0.706 &  0.896 & 1.82G  &  11.78 M  \\
        \modelres       & 0.707 &  0.895 &  1.8207G     & 11.96M  \\
        \noshare-cmp    & 0.698 &  0.893 &1.30G & 7.51M \\
        \noshare    & $\bm{0.710}$  &   $\bm{0.899}$  & 1.67G  & 10.69M \\

    \hline
    \end{tabular}
\label{tab:nosharing_resnet_imagenet}
\end{table*}

\subsection{Image classification with non-local blocks}
\label{ssec:nosharing_experiments_image_nonlocal}

In this section, an experimental comparison and validation is conducted with non-local blocks. The benchmark of CIFAR100 is selected, while \resnet18 is referred as the baseline. The original non-local network (NL) and the disentangled non-local (DNL) are the main compared methods from the literature. As a reminder we perform two extensions in DNL: i) we add a first-degree term (\noshare-$NL^{[3]}$), ii) we add a fourth degree term (\noshare-$NL^{[4]}$).

Table~\ref{tab:nosharing_resnet_nonlocal} contains the accuracy of each method. Notice that DNL improves upon the baseline NL, while \noshare-$NL^{[3]}$ improves upon both DNL and NL. Interestingly, the fourth-degree variant, i.e., \noshare-$NL^{[4]}$ outperforms all the compared methods by a considerable margin without increasing the number of parameters significantly. The results verify our intuition that the different polynomial terms enable additional flexibility to the model. 

Besides the experiments on CIFAR100, we also test the proposed Non-local blocks in ImageNet. The training setup remains the same as in sec.~\ref{ssec:nosharing_experiments_image_resnet}. As shown in Table~\ref{tab:nosharing_resnet_nonlocal_imagenet}, the proposed Non-local blocks with different polynomial degrees significantly improve the top-1 accuracy on ImageNet while the total parameter numbers are lower than the baseline \resnet18. 
More specifically, we set the bottleneck ratio of $4$ for the proposed \noshare-$NL^{[3]}$ and \noshare-$NL^{[4]}$, and we apply non-local blocks in multiple layers (c3+c4+c5) to better capture long-range dependency with only a slight increase in the computation cost. \noshare-$NL^{[4]}$ significantly outperforms NL by $1.03\%$, showing the advantages of the polynomial information fusion.

\begin{table}[h]
\centering
    \begin{minipage}{.48\linewidth}
        \caption{Classification on CIFAR100 with non-local blocks.}
         \begin{tabular}{|c | c | c|}
             \hline
            \textbf{ Model} & \textbf{\#p} $\times 10^6$ & \textbf{Accuracy}\\
            \hline
             \resnet18      & $11.2$ & $0.756$\\\hline
             NL             & $11.57$ & $0.769$ \\\hline
             DNL            & $11.57$ & $0.771$ \\\hline
             \noshare-$NL^{[3]}$    & $11.87$ & ${0.773}$\\\hline
             \noshare-$NL^{[4]}$    & $12.00$ & $\bm{0.779}$\\\hline
         \end{tabular}
        \label{tab:nosharing_resnet_nonlocal}
    \end{minipage}
    \begin{minipage}{.48\linewidth}
        \caption{Classification on ImageNet with non-local blocks.}
         \begin{tabular}{|c | c | c|}
             \hline
            \textbf{Model} & \textbf{\#p} $\times 10^6$ & \textbf{Accuracy}\\
            \hline
             \resnet18      & $11.69$ & $0.698$\\\hline
             NL             & $12.02$ & $0.702$ \\\hline
             PDC   & $10.69$ & $0.710$ \\\hline
             PDC-$NL^{[3]}$    & $11.35$ & $0.712$\\\hline
             PDC-$NL^{[4]}$    & $11.51$ & $\bm{0.716}$\\\hline
         \end{tabular}
        \label{tab:nosharing_resnet_nonlocal_imagenet}
    \end{minipage}
\end{table}

\subsection{Audio classification}
\label{ssec:nosharing_experiments_audio}

Besides image classification, we conduct a series of experiments on audio classification to test the generalization of the polynomial expansion in different types of signals. The popular dataset of Speech Commands~\cite{warden2018speech} is selected as our benchmark. The dataset consists of $60,000$ audio files containing a single word each. The total number of words is $35$, while there are at least $1,500$ different files for each word. Every audio file is converted into a mel-spectrogram. 

\resnet7 includes one residual block per group, while \resnet18 includes two residual blocks per group. 
The accuracy for each model (in both \resnet7 and \resnet18 comparisons) is reported in Table~\ref{tab:nosharing_resnet_speech_command}.  All the compared methods have accuracy over $0.97$, while the polynomial expansions of \modelres{} and \noshare{} are able to reduce the number of parameters required to achieve the same accuracy, due to their expressiveness.

\begin{table}[h]
\centering
    \caption{Speech classification with \resnet{} variants. Four residual blocks are used in \resnet7 instead of eight of \resnet18. Nevertheless, the respective \noshare7 can reduce even further the parameters to achieve the same performance.}
     \begin{tabular}{|c | c | c|}
         \hline
         \textbf{Model} & \textbf{\# param ($\times 10^6$)} & \textbf{Accuracy}\\
        \hline
        \resnet7           & $4.9$ & $0.974$\\\hline
         \sne              & $5.1$ & $0.974$\\\hline
         \noshare          &  $\bm{3.9}$ & $\bm{0.975}$\\\hline \hline
         
         \resnet18           & $11.2$ & $0.977$\\\hline
         \sne               & $11.5$ & $0.977$\\\hline
         \modelres          &  $\bm{6.0}$ & $0.977$\\\hline
         \noshare           & $8.0$ & $\bm{0.978}$\\ \hline
     \end{tabular}
     
 \label{tab:nosharing_resnet_speech_command}
\end{table}

\subsection{Image classification on long-tailed distributions}
\label{ssec:nosharing_experiments_long_tailed_distribution}

We scrutinize the performance of the proposed models on long-tailed image recognition. We convert CIFAR10, which has $5,000$ number of samples per class, to a long-tailed version, called CIFAR10-LT. The imbalance factor (IF) is defined as the ratio of the largest class to the smallest class. The imbalance factor varies from $10-200$, following similar benchmarks tailored to long-tailed distributions~\citep{cui2019class}. We note that the models are as defined above; the only change is the data distribution. The accuracy of each method is reported in Table~\ref{tab:nosharing_resnet_cifar10_long_tailed}. The results exhibit the benefits of the proposed models, which outperform the baselines.

\begin{table}[htb]
\centering
    \caption{Accuracy on image classification on CIFAR10-LT. Each column corresponds to a different imbalance factor (IF). }
     \begin{tabular}{|c | c | c| c | c | c | }
         \hline
        \backslashbox{Model}{IF}&\makebox[3em]{200}&\makebox[3em]{100}&\makebox[3em]{50}&\makebox[3em]{20}&\makebox[3em]{10}\\\hline\hline
        \hline
         \resnet18          & $0.645$ & $0.696$ & $0.784$ & $0.844$ & $0.877$\\\hline
         \sne               & $0.636$ & $0.713$ & $0.784$ & $0.844$ & $0.878$\\\hline
         \modelres          & $0.653$ & $0.718$ & $0.783$ & $0.845$ & $0.879$\\\hline
         \noshare-comp      & $0.653$ & $\bm{0.727}$ & $0.786$ & $0.848$ & $0.882$\\\hline
         \noshare-param     & $\bm{0.665}$ & $0.726$ & $\bm{0.792}$ & $\bm{0.851}$ & $\bm{0.886}$\\\hline
     \end{tabular}
 \label{tab:nosharing_resnet_cifar10_long_tailed}
\end{table}

 \section{Related work}
\label{sec:nosharing_related}

Classification benchmarks~\cite{russakovsky2015imagenet} have a profound impact in the progress observed in machine learning. Such benchmarks been a significant testbed for new architectures~\cite{krizhevsky2012imagenet, he2015deep, hu2018squeeze, he2015convolutional} and for introducing novel tasks, such as adversarial perturbations~\cite{szegedy2013intriguing}. The architectures originally designed for classification have been applied to diverse tasks, such as image generation~\cite{brock2019large, zhang2018self}.

\resnet~\cite{he2015deep} has been among the most influential works of the last few years which can be attributed both in its simplicity and its stellar performance. \resnet{} has been studied in both its theoretical capacity~\cite{hardt2016identity, balduzzi2017shattered, shamir2018resnets, zaeemzadeh2018norm} and its empirical performance~\cite{xie2017aggregated, szegedy2017inception, zagoruyko2016wide}.  A number of works focus on modifying the shortcut connection and the concatenation operation~\cite{huang2017densely, chen2017dual, wang2018mixed}. 

The Squeeze-and-Excitation network (\sne)~\cite{hu2018squeeze} has been extended by \citet{roy2018concurrent} to capture second-degree correlations in both the spatial and the channel dimensions. \cite{hu2018gather} extend \sne{} by replacing the pooling operation with alternative operators that aggregate contextual information. \citet{yang2020gated} inspired by the \sne{} propose a gated convolution module. \citet{ruan2020linear} improve \sne{} by introducing long-range dependencies. \sne{} has also been used as a drop-in module to improve the performance of residual blocks~\cite{gao2019res2net}.

Non-local neural networks have been used extensively for capturing long-range interactions in both image-related tasks~\cite{li2020neural, zhu2019asymmetric, huang2019ccnet} and video-related tasks~\cite{chen20182, girdhar2019video}. Non-local networks are also related with self-attention~\cite{vaswani2017attention} that is widely used in both vision~\cite{parmar2019stand} and natural language processing~\cite{galassi2019attention, ma2019tensorized}. 
\citet{cao2019gcnet} study when the long-range interactions emerge in non-local neural networks. They also frame a simplified non-local block, which in our terminology is a second-degree polynomial. Naturally, they observe that this resembles the \sne{} and merge their simplified non-local block and \sne{} block.

A promising line of research is that of polynomial activation functions. For instance, the element-wise quadratic activation function $f^2$ applied to a linear operation $\bm{C}\mvinvar$ is $(\bm{C}\mvinvar)^2$. Both the theoretical~\cite{kileel2019expressive, livni2014computational} and the empirical results~\cite{ramachandran2017searching, lokhande2020generating} support that polynomial activation functions can be beneficial. Our work is orthogonal to the polynomial activation functions, as they express a polynomial expansion of the representation, while we model a polynomial expansion of the input.

A well-established line of research is that of considering second or higher-order moments for various tasks. For instance, second-order statistics have been used for normalization methods~\cite{huang2018decorrelated}, learning latent variable models~\cite{anandkumar2014tensor}. However, our work focuses on classification methods that can be expresses as polynomials and not on the method of moments. 

Tensors and tensor decompositions are related to our work~\citep{Sidiropoulos:16}. 
Tensor decompositions such as the CP or the Tucker decompositions~\citep{kolda2009tensor} are frequently used for model compression in computer vision. Tensor decompositions have also been used for modeling the components of deep neural networks. \citet{cohen2016convolutional} interpret a whole convolutional network as a tensor decomposition, while the recent Einconv~\citep{hayashi2019einconv} focus on a single convolutional layer and model them with tensor decompositions. In our work the focus is not in the tensor decomposition used, but on the polynomial expansion that provides insights on the correlations that are captured by each model.

A line of research that is related to ours is that of multiplicative data fusion~\cite{jayakumar2020Multiplicative, markos, markos2, chrysos2021conditional, reed2014learning, yu2017multi, kim2018bilinear}. 
Even though multiplicative interactions can be considered as second-degree polynomials, data fusion of the aforementioned works is not our focus.

 \section{Conclusion}

In this work, we study popular classification networks under the unifying perspective of polynomials. Notably, the popular \resnet, \sne{} and non-local networks are expressed as first, second and third degree polynomials respectively. The common framework provides insights on the inductive biases of each model and enables natural extensions building upon their polynomial nature. We conduct an extensive evaluation on image and audio classification benchmarks. 
We show how intuitive extensions to existing networks, e.g., converting the third-degree non-local network into a fourth degree, can improve the performance.
Such natural extensions can be used for designing new architectures based on the proposed taxonomy.  
Importantly, our experimental evaluation highlights the dual utility of the polynomial framework: the networks can be used either for model compression or increased model performance. We expect this to be a significant feature when designing architectures for edge devices. Our experimentation in the presence of limited data and long-tailed data distributions highlights the benefits of the proposed taxonomy and provides a link to real-world applications, where massive data annotation is challenging.

\clearpage
\bibliographystyle{splncs04}
\bibliography{egbib}

\newpage
\appendix

\section{Detailed notation}
\label{sec:nosharing_notation_supplem}

\textbf{Products}: The \textit{Hadamard} product of $\bm{A}, \bm{B} \in \realnum^{I \times N}$ is defined as $\bm{A} * \bm{B}$ and is equal to ${a}_{(i, j)} {b}_{(i, j)}$ for the $(i, j)$ element. The \textit{Khatri-Rao} product of matrices $\bm{A} \in \realnum^{I \times N}$
and $\bm{B} \in \realnum^{J \times N}$ is
denoted by $\bm{A} \odot \bm{B}$ and yields a matrix of
dimensions $(IJ)\times N$.  The Khatri-Rao product for a set of matrices  $\{\bm{A}\matnot{m} \in \realnum^{I_m \times N} \}_{m=1}^M$ is abbreviated by $\bm{A}\matnot{1} \odot \bm{A}\matnot{2} \odot  \cdots \odot  \bm{A}\matnot{M} \doteq  \bigodot_{m=1}^M \bm{A}\matnot{m}$.

\textbf{Tensors}: Each element of an $M^{th}$ order tensor $\bmcal{X}$ is addressed by $M$ indices, i.e., $(\bmcal{X})_{i_{1}, i_{2}, \ldots, i_{M}} \doteq x_{i_{1}, i_{2}, \ldots, i_{M}}$. An $M^{th}$-order tensor $\bmcal{X}$ is  defined over the
tensor space $\realnum^{I_{1} \times I_{2} \times \cdots \times
I_{M}}$, where $I_{m} \in \mathbb{Z}$ for $m=1,2,\ldots,M$. 
 The \textit{mode-$m$ unfolding} of a tensor $\bmcal{X} \in
 \realnum^{I_1 \times I_2 \times \cdots \times I_M}$ maps
 $\bmcal{X}$ to a matrix $\bm{X}_{(m)} \in \realnum^{I_{m}
 \times \bar{I}_{m}}$ with $\bar{I}_{m}= \prod_{k=1 \atop k  \neq m}^M I_k $ such
 that the tensor element $x_{i_1, i_2, \ldots, i_M}$ is
 mapped to the matrix element $x_{i_{m}, j}$ where
 $j=1 + \sum_{k=1 \atop k \neq m}^M (i_k - 1) J_k$ with $J_k =
\prod_{n =1 \atop n \neq m}^{k-1} I_n $. 
The \textit{mode-$m$ vector product} of $\bmcal{X}$ with a
vector $\bm{c} \in \realnum^{I_m}$, denoted by
$\bmcal{X} \times_{m} \bm{c} \in \realnum^{I_{1}\times
I_{2}\times\cdots\times I_{m-1}  \times I_{m+1} \times
\cdots \times I_{M}} $, results in a tensor of order $M-1$:
\begin{equation}\label{E:Tensor_Mode_n}
(\bmcal{X} \times_{m} \bm{c})_{i_1, \ldots, i_{m-1}, i_{m+1},
\ldots, i_{M}} = \sum_{i_m=1}^{I_m} x_{i_1, i_2, \ldots, i_{M}} u_{i_m}.
\end{equation}
The \textit{CP decomposition}~\citep{kolda2009tensor} factorizes a tensor into a sum of component rank-one tensors. The rank-$R$ CP decomposition of an $M^{th}$-order tensor $\bmcal{X}$ is written as:
 \begin{equation}\label{E:CP}
\bmcal{X}  \doteq [\![ \bm{C}\matnot{1}, \bm{C}\matnot{2}, \ldots, \bm{C}\matnot{M}  ]\!] =  \sum_{r=1}^R \bm{c}_r^{(1)}  \circ \bm{c}_r^{(2)}  \circ \cdots \circ \bm{c}_r^{(M)},
\end{equation}
where $\circ$ is the vector outer product. The factor matrices $\big\{ \bm{C}\matnot{m} = [\bm{c}_1^{(m)},\bm{c}_2^{(m)}, \cdots, \bm{c}_R^{(m)} ] \in \mathbb{R}^{I_m \times R} \big\}_{m=1}^{M}$ collect 
the vectors from the rank-one components. By considering the mode-$1$ unfolding of $\bmcal{X}$, the CP decomposition can be written in matrix form as: 

 \begin{equation}
 \label{eq:polygan_cp_unfolding}
\bm{X}_{(1)}  
\doteq \bm{C}\matnot{1} \bigg( \bigodot_{m = M}^{2} \bm{C}\matnot{m}\bigg)^T
\end{equation}

The following lemma is useful in our method: 

\begin{lemma} [\citet{chrysos2019polygan}]
For a set of $N$ matrices $\{\bm{A}\matnot{\nu} \in \realnum^{I_{\nu} \times K} \}_{\nu=1}^N$  and $\{\bm{B}\matnot{\nu} \in \realnum^{I_{\nu} \times L} \}_{\nu=1}^N$, the following equality holds:
\begin{equation}
    (\bigodot_{\nu=1}^N \bm{A}\matnot{\nu})^T \cdot (\bigodot_{\nu=1}^N \bm{B}\matnot{\nu}) = (\bm{A}\matnot{1}^T \cdot \bm{B}\matnot{1}) * \ldots * (\bm{A}\matnot{N}^T \cdot \bm{B}\matnot{N})
\label{eq:polygan_suppl_lemma1_N}
\end{equation}
\label{lemma:polygan_lemma_hadamard_kr2}
\end{lemma}

\section{Polynomials as a single tensor product}
\label{ssec:nosharing_explanation_motivational_figure}

As mentioned in the main paper polynomials and tensors are closely related. To illustrate the differences between the proposed variant of sec.~\ref{ssec:nosharing_pi_net_nosharing} and the proposed taxonomy we can formulate them as a single tensor product. We assume a second-degree polynomial expansion of \eqref{eq:nosharing_elementwise_poly}. The tensors are then up to third-order, which enables a visualization (as in the Fig.\ref{fig:graphical_abstract}). The initial equation is: 
\begin{equation}
    \bm{y} = \bm{\beta} + \Big(\bm{W}^{[1]}\Big)^T \binvar + \bigg(\bmcal{W}^{[2]} \times_{2} \binvar \times_{3} \binvar \bigg)
    \label{eq:approx_second_order_tensor}
\end{equation}

The $\tau^{th}$ output of \eqref{eq:approx_second_order_tensor} can be written in element-wise form as:
\begin{equation}
    y_{\tau} = {\beta}_{\tau} + \sum_{k=1}^{\delta} w^{[1]}_{\tau, k} \invar_k + \sum_{k, m = 1}^{\delta} w^{[2]}_{\tau, k, m} \invar_k \invar_m
    \label{eq:approx_second_order_elementwise}
\end{equation}

We can collect all the parameters of \eqref{eq:approx_second_order_tensor} under a single tensor by padding the input $\binvar \in \realnum^{\delta}$. Specifically, if we consider the padded version $\tilde{\binvar} = [\invar_1, \dots, \invar_{\delta}, 1]^T$, then \eqref{eq:approx_second_order_tensor} can be written in the format $\bm{y} = \tilde{\bmcal{W}}\times_{2} \tilde{\binvar} \times_{3} \tilde{\binvar}$ as we demonstrate below. 

The $\tau^{th}$ output of  $\tilde{\bmcal{W}}\times_{2} \tilde{\binvar} \times_{3} \tilde{\binvar}$ is:
\begin{equation}
\centering
    \begin{split}
        y_{\tau} = \sum_{k, m=1}^{\delta + 1} \tilde{w}_{\tau, k, m} \tilde{\invar}_k \tilde{\invar}_m = \underbrace{\tilde{w}_{\tau, \delta + 1, \delta + 1}}_{\text{constant term}} + \\
        \underbrace{\sum_{m=1}^{\delta} \tilde{w}_{\tau, \delta + 1, m} \invar_m + \sum_{k=1}^{\delta} \tilde{w}_{\tau, k, \delta + 1} \invar_k}_{\text{first-degree term}} + 
        \underbrace{\sum_{k, m = 1}^{\delta} \tilde{w}_{\tau, k, m} \invar_k \invar_m}_{\text{second-degree term}}
    \end{split}
    \label{eq:approx_second_order_elementwise_single_tensor}
\end{equation}

If we set: 

\begin{equation}
    \begin{cases}
        \beta_{\tau} =  \tilde{w}_{\tau, \delta + 1, \delta + 1} & \\ 
        w^{[1]}_{\tau, k} =  \tilde{w}_{\tau, \delta + 1,k } + \tilde{w}_{\tau, k, \delta + 1}  & \text{for } k = 1, \dots, \delta \\ 
        w^{[2]}_{\tau, k, m} =  \tilde{w}_{\tau, k, m}  & \text{for } k, m = 1, \dots, \delta \\ 
    \end{cases}
\end{equation}

then \eqref{eq:approx_second_order_elementwise_single_tensor} becomes the polynomial expansion of \eqref{eq:approx_second_order_elementwise}. 

This enables us to express different degree polynomial expansions with a third-order tensor. The first-degree methods, e.g., \resnet~\cite{he2015deep}, have $w^{[2]}_{\tau, k, m}=0$, while \sne~\cite{hu2018squeeze} assumes $w^{[1]}_{\tau, k}=0$. The ${\Pi}$-net family assumes low-rank decomposition with shared factors, i.e., the low-rank decompositions of ${\bmcal{W}^{[n]}}_{n=1}^{N}$ share factor matrices. On the contrary, our proposed \noshare{} does not assume a sharing pattern, thus it can express independently the terms $\bm{W}^{[1]}, \bmcal{W}^{[2]}$.

\section{Proofs}
\begin{claim}
    The Squeeze-and-excitation block of \eqref{eq:nosharing_senet_block} is a special form of second-degree polynomial term. 
\end{claim}

\begin{proof}
    The global pooling function on a matrix $\bm{C}$ can be expressed as $\frac{1}{hw}\overrightarrow{\bm{1}}^T \bm{C}$. The $r$ function that replicates the channels acts on a vector $\bm{c}$ and results in the expression $\overrightarrow{\bm{1}}\bm{c}^T$. 
    
    The identity  $\bm{X} * \bm{a}\bm{b}^T = diag(\bm{a}) \bm{X} diag(\bm{b})$ can be used to convert the Hadamard product of \eqref{eq:nosharing_senet_block} into a matrix multiplication \citep{styan1973hadamard}.  Then, \eqref{eq:nosharing_senet_block} becomes:
    \begin{align}
    \centering
    \begin{split}
        {\bm{Y}}_{s} = (\minvar\bm{C}_1) * \overrightarrow{\bm{1}} \bigg(\Big(\frac{1}{hw}\overrightarrow{\bm{1}}^T\minvar\bm{C}_1\Big)\bm{C}_2\bigg)^T = 
        (\minvar\bm{C}_1) \frac{1}{hw}diag(\bm{C}_2^T\bm{C}_1^T \minvar^T \overrightarrow{\bm{1}}) = \\
        (\minvar\bm{C}_1) \frac{1}{hw} \bm{\mathcal{I}} \times_3 (\bm{C}_2^T\bm{C}_1^T \minvar^T \overrightarrow{\bm{1}})
    \end{split}
    \end{align}
    where as a reminder $\bm{\mathcal{I}}$ is a third-order super-diagonal unit tensor. The last equation is a second-degree term with $\matra{1}{2}(\minvar) = \minvar\bm{C}_1$ and $\matra{2}{2}(\minvar) = \frac{1}{hw} \bm{\mathcal{I}} \times_3 (\bm{C}_2^T\bm{C}_1^T \minvar^T \overrightarrow{\bm{1}})$.
\end{proof}

\section{Auxiliary experiments}

\begin{table}[h]
\centering
    \caption{Image classification on CIFAR100 with variants of \resnet34.} 
     \begin{tabular}{|c | c | c|}
         \hline
         \textbf{Model} & \textbf{\# param ($\times 10^6$)} & \textbf{Accuracy}\\
        \hline
         \resnet34 & $21.3$ & $0.769$\\\hline
         \modelres &  ${14.7}$ & $0.769$\\\hline
         \noshare-channels & $36.3$ & $\bm{0.774}$\\\hline  
         \noshare &  $\bm{10.5}$ & $0.770$\\\hline
     \end{tabular}
     
 \label{tab:nosharing_resnet_cifar100_resnet34}
\end{table}

\subsection{Image classification with limited data}
\label{ssec:nosharing_experiments_limited_data}

\begin{figure}
\centering
    \centering
    \includegraphics[width=0.9\linewidth]{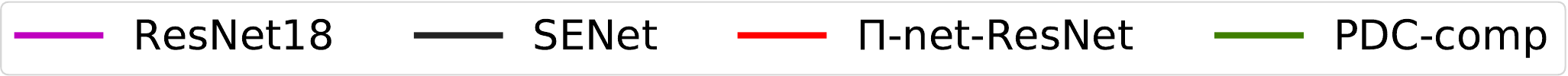} \\
    \includegraphics[width=0.94\linewidth]{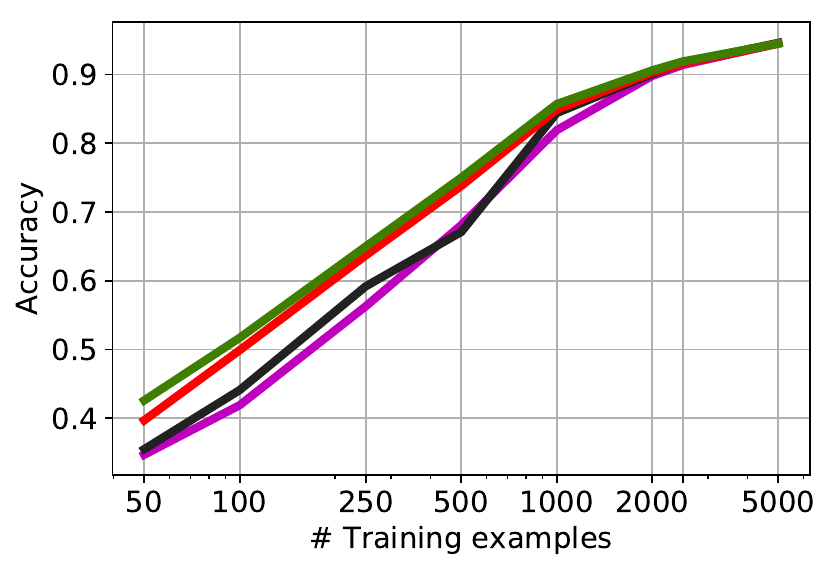}
\caption{Image classification with limited data. The x-axis declares the number of training samples per class (log-axis). As the number of samples is reduced (i.e., moving from right to the left), the performance gap between \modelres{} and \resnet18 increases. Similarly, \noshare-comp performs better than \modelres, especially in the limited data regimes on the left.}
\label{fig:nosharing_experiment_limited_data}
\end{figure}

A number of experiments is performed by progressively reducing the number of training samples per class. The number of samples is reduced uniformly from the original $5,000$ down to $50$ per class, i.e., a $100\times$ reduction, in CIFAR10. The architectures of Table~\ref{tab:nosharing_resnet_cifar10} (similar to \resnet18) are used unchanged; only the number of training samples is progressively reduced. The resulting Fig.~\ref{fig:nosharing_experiment_limited_data} visualizes the performance as we decrease the training samples. The accuracy of \resnet18 decreases fast for limited training samples. \sne{} deteriorates at a slower pace, steadily increasing the difference from \resnet18 (note that both share similar number of parameters). \modelres{} improves upon \sne{} and performs better even under limited data. However, the proposed \noshare-comp outperforms all the compared methods for $50$ training samples per class. The difference in the accuracy between \noshare{} and \modelres{} increases as we reduce the number of training samples. Indicatively, with $50$ samples per class, \resnet18 attains accuracy of $0.347$, \sne{} scores $0.355$, \modelres{} scores $0.397$ and \noshare-comp scores $0.426$, which is a $22\%$ increase over the \resnet18 baseline.

\subsection{Classification without activation functions}

Typical feed-forward neural networks, such as CNNs, require activation functions to learn complex functions~\cite{hornik1989multilayer}. However, the proposed view of polynomial expansion enables capturing higher-order correlations even in the absence of activation functions.
That is, the expressivity of higher-degree polynomials can be assessed without activation functions. We conduct a series of experiments on all three datasets with higher-degree polynomials. Our core experiments study the higher-degree polynomials of $\Pi$-nets~\cite{chrysos2020poly}, versus the proposed model of \eqref{eq:nosharing_model_no_sharing}. We also implement the \resnet{} without activation functions to assess how first-degree polynomials perform. 

For the first experiment, we utilize \resnet18 as the backbone and test the baselines on CIFAR100. Three variations of $Pi-$net are considered as the compared methods: one with second-degree, one with third-degree and one with fourth-degree residual blocks. The same polynomial expansions are used for the proposed \noshare. 
The accuracy of each method is reported in Table~\ref{tab:nosharing_resnet_cifar100_linear}. All the variants of \modelres{} and \noshare{} exhibit a high accuracy based solely on the high-degree polynomial expansion. However, \modelres{} saturates when the residual block is a third or fourth degree polynomial, while the \noshare{} does not suffer from the same issue. On the contrary, the performance of the \noshare{} variant with third and forth degree residual block outperforms the second-degree residual block. 

\begin{table}[h]
\centering
    \caption{Image classification on CIFAR100 without activation functions. Both \modelres{} and \noshare{} use high-degree polynomial expansion to achieve high accuracy even in the absence of activation functions. The proposed \noshare{} achieves both increased performance and improves its performance when each residual block has third or fourth degree polynomial instead of second.}
     \begin{tabular}{|c | c | c|}
         \hline
         \textbf{Model} & \textbf{\# param ($\times 10^6$)} & \textbf{Accuracy}\\
        \hline
         \resnet18 & $11.2$ & $0.168$\\\hline
         \modelres &  $11.9$ & $0.667$\\\hline
         \modelres$^{[3]}$ &  $11.2$ & $0.648$\\\hline
         \modelres$^{[4]}$ &  $11.2$ & $0.626$\\\hline
         \noshare &  $\bm{5.46}$ & $0.689$\\\hline
         \noshare$^{[3]}$ &  $11.2$ & $\bm{0.703}$\\\hline
         \noshare$^{[4]}$ &  $18.8$ & $0.699$\\\hline
     \end{tabular}
 \label{tab:nosharing_resnet_cifar100_linear}
\end{table}

The models are also evaluated on CIFAR10 with \resnet18 and three variants of $\Pi-$nets as the backbone. Three variants of \noshare{} with different expansion degrees are designed. The results are tabulated on Table~\ref{tab:nosharing_resnet_cifar10_linear}. Each variant of \modelres{} and \noshare{} surpasses the $0.87$ accuracy and outperform the \resnet18 by a wide margin. 
In contrast to \modelres{}, the performance of \noshare{} does not decrease when the degree of the residual block increases, \ie{} from second to fourth-degree. Overall, \noshare{} outperforms $\Pi-$net.

\begin{table}[h]
\centering
    \caption{Image classification on CIFAR10 without activation functions. The results illustrate the expressiveness of the proposed model even in the absence of activation functions. Notice that \noshare$^{[3]}$ improves upon \noshare{} with second-degree blocks. On the contrary, this does not happen to the compared \modelres.}
     \begin{tabular}{|c | c | c|}
         \hline
         \textbf{Model} & \textbf{\# param ($\times 10^6$)} & \textbf{Accuracy}\\
        \hline
         \resnet18 & $11.2$ & $0.391$\\\hline
         \modelres &  $11.9$ & $0.907$\\\hline
         \modelres$^{[3]}$ &  $11.2$ & $0.891$\\\hline
         \modelres$^{[4]}$ &  $11.2$ & $0.877$\\\hline
         \noshare &  $\bm{5.4}$ & $0.909$\\\hline
         \noshare$^{[3]}$ &  $11.2$ & $\bm{0.918}$\\\hline
         \noshare$^{[4]}$ &  $18.8$ & $\bm{0.918}$\\\hline
     \end{tabular}
 \label{tab:nosharing_resnet_cifar10_linear}
\end{table}

The last experiment is conducted on the Speech Commands dataset. The baseline of \resnet18 is selected, while the \modelres{} is the compared method. The results in Table~\ref{tab:nosharing_resnet_speech_linear} depict the same motif: the two polynomial expansions are very expressive. Impressively, in this dataset the result without activation functions is only $0.007$ decreased when compared to the respective results with activation functions. This highlights that simple datasets might not always demand activation functions to achieve high-accuracy.

\begin{table}[h]
\centering
    \caption{Audio classification without activation functions.}
     \begin{tabular}{|c | c | c|}
         \hline
         \textbf{Model} &\textbf{ \# param ($\times 10^6$)} & \textbf{Accuracy}\\
        \hline
         \resnet18 & $11.2$ & $0.464$\\\hline
         \modelres &  $11.9$ & $0.971$\\\hline
         \noshare &  $\bm{5.4}$ & $\bm{0.972}$\\\hline
     \end{tabular}
 \label{tab:nosharing_resnet_speech_linear}
\end{table}

\begin{table}[ht!]
\caption[caption]{\textbf{COCO object detection and segmentation results} using Mask-RCNN and Cascade Mask-RCNN. The backbone models are pre-trained ResNet18 and PDC-ResNet18 models on ImageNet-1K. We employ MMDetection with $1\times$ schedule.
\label{tab:grigorisrebuttal2}
}
\tablestyle{6pt}{1.1}
\addtolength{\tabcolsep}{-4.5pt}
\vspace{2ex}
\begin{tabular}{@{}lcccccc@{}}
\hline
backbone & $\text{AP}^{\text{box}}$ & $\text{AP}^{\text{box}}_{50}$ & $\text{AP}^{\text{box}}_{75}$ & $\text{AP}^{\text{mask}}$ & $\text{AP}^{\text{mask}}_{\text{50}}$ & $\text{AP}^{\text{mask}}_{75}$  \\
\hline
\multicolumn{7}{c}{{Mask-RCNN 1$\times$ schedule}} \\
ResNet18      & 33.9 & 53.9 &36.2 &31.0 & 50.9 & 33.0 \\
\gr
PDC-ResNet18  & 34.8 & 55.2 & 37.4 & 31.8 & 52.2 & 34.1 \\
\hline
\multicolumn{7}{c}{{Cascade Mask-RCNN 1$\times$ schedule}} \\
ResNet18        & 37.3 &54.8 &40.4 &32.6 &52.2 &34.9 \\
\gr
PDC-ResNet18    & 38.1 & 55.9 & 41.7 & 33.2 & 53.3 & 35.7 \\
\hline
\\
\end{tabular}

\end{table}

\subsection{Object detection and segmentation}

We adopt MS COCO 2017 \cite{Lin2014} as the primary benchmark for the experiments of object detection and segmentation. We use the train split (118k images) for training and report the performance on the val split (5k images). We employ standard evaluation metrics for COCO dataset, where multiple IoU thresholds from 0.5 to 0.95 are applied. The detection results are evaluated with mAP.

We use the final model weights
from ImageNet-1K pre-training as network initializations and fine-tune Mask R-CNN~\cite{He2017} and Cascade Mask R-CNN~\cite{Cai2018} on the COCO dataset. Following default settings in MMDetection, we use the 1$\times$ schedule (\ie 12 epochs). 

Table~\ref{tab:grigorisrebuttal2} shows object detection and instance segmentation results comparing ResNet18 and the proposed PDC-ResNet18. As we can see from the results, the proposed PDC-ResNet18 achieves an obvious better performance than the baseline ResNet18 in terms of the box and mask AP, confirming the effectiveness of the proposed polynomial learning scheme.

\begin{figure} 
\centering
\centering
\includegraphics[width=0.45\linewidth]{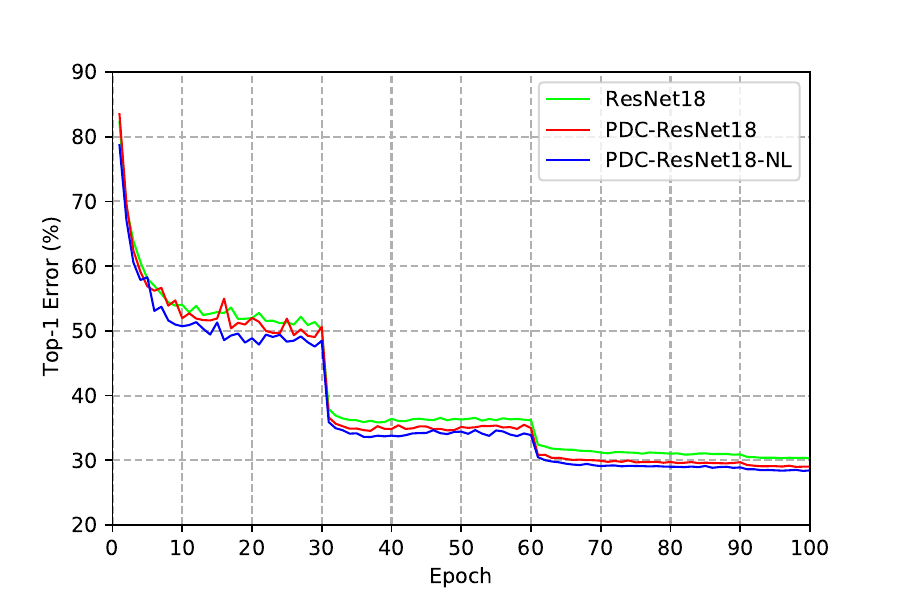}
\caption{Top-1 validation error on ImageNet with proposed PDC and NL methods throughout the training.}
\label{fig:imagenet}
\end{figure}

\end{document}